\documentclass{article}
\usepackage{graphicx}
\usepackage[english]{babel}
\usepackage{amsfonts} 
\usepackage{xparse}
\usepackage{comment}
\usepackage[normalem]{ulem}
\usepackage{diagbox} 
\usepackage[utf8]{inputenc}
\usepackage{makecell} 
\usepackage{xcolor}
\usepackage[shortlabels]{enumitem}
\usepackage{bm}
\usepackage{authblk}
\usepackage[letterpaper,top=2cm,bottom=2cm,left=2.5cm,right=4.5cm,marginparwidth=3.5cm]{geometry}
\usepackage{subcaption}
\usepackage[numbers,compress]{natbib}
\usepackage{booktabs}
\usepackage{amsmath}
\usepackage{amsthm}
\usepackage{soul} 
\usepackage{url} 

\usepackage{amssymb}

\newcommand{\gptA}{\texttt{gpt-4.1}}
\newcommand{\gptB}{\texttt{gpt-5.6-sol}}
\newcommand{\opus}{\texttt{claude-opus-5}}
\newcommand{\sonnet}{\texttt{claude-sonnet-5}}
\newcommand{\gem}{\texttt{gemini-3.7-flash}}
\newcommand{\sonnetold}{\texttt{claude-sonnet-4-5}}

\title{Measuring Model-Specific Alignment Drift \\ in LLM-Based Opinion Dynamics}
 
\author[a]{Nata\v sa Djurdjevac Conrad}
\author[a,b]{Sebastian Pokutta}
\author[a,c]{Christof Sch\"utte}
\affil[a]{Zuse Institute Berlin, Germany}
\affil[b]{Institut f\"ur Mathematik, Technische Universität Berlin, Germany}
\affil[c]{Institut f\"ur Mathematik und Informatik, Freie Universit\"at Berlin, Germany}

\begin{document}

\maketitle

\begin{abstract}
Large language model (LLM) agents are increasingly used as synthetic participants in
opinion-dynamics experiments, often assuming that an agent's behaviour is fully determined by its
prompt. We test this assumption with a framework in which LLM agents discuss a contested policy issue
over repeated rounds while their stance is quantified by a fixed psychometric instrument
independent of any dynamics model. Across six LLMs, committed personas are intrinsically stable: in each of the five models tested
without peer input, every agent keeps its assigned side for 20 rounds. The observed loss of alignment under discussion is therefore interaction-induced---we call it
\emph{alignment drift}---and model-specific: \texttt{gpt-4.1} keeps opinion clusters apart, 
\texttt{claude-sonnet-5} holds them partially, and \texttt{gpt-5.6-sol} and
\texttt{claude-opus-5} abandon them. Under selective attention, one LLM 
maintains separated opinion clusters
under stance-homophilous attention. Qualitative profiles persist at doubled
population and horizon and on a second issue. Such heterogeneity makes the LLM choice a first-class validity
variable for social simulation.
\end{abstract}

\section{Introduction}
LLMs are becoming a foundation for computational social science: instantiated as agents with
personas, they argue, persuade, and update positions in simulated discussions, promising
synthetic public opinion at scale
\citep{park2023generative,argyle2023out,horton2023large,anthis2025llm,bail2024can,ziems2024can,haase2026staticresponsesmultiagentllm}.
A common premise is that an agent's behaviour is controlled by its prompt, making the LLM
interchangeable. Evidence questions this assumption: LLMs carry their own opinion distributions
\citep{santurkar2023whose}, exhibit sycophancy \citep{sharma2023understanding}, and produce
biased simulations \citep{qu2024performance}; agent populations converge
\citep{chuang2024simulating,cau2025lodas}, form echo chambers
\citep{wang2024echo,donkers2025understanding}, develop collective biases
\citep{ashery2025emergent}, and show judgment asymmetries \citep{loru2025simulation}. Yet the
evidence is largely qualitative: formal opinion models are rarely fitted to LLM trajectories
\citep{he2026dialog}, and no prior study performs model selection across alternative dynamics
models while varying the LLM under a matched protocol---the precondition for attributing
simulated dynamics to the LLM rather than the experimental setting.

\subsection{Related work}
\textbf{LLM-based opinion dynamics.} LLM-agent networks reproduce consensus, echo chambers,
and polarization under broadcast, networked, and recommender-like exposure
\citep{chuang2024simulating,wang2024echo,donkers2025understanding,torneberg2023simulating,
yang2024oasis}; \emph{selective agreement} has been identified as the micro-level mechanism
\citep{cau2025lodas,cau2025selective}. Only \citet{he2026dialog} fit
formal opinion models to LLM trajectories; none compares multiple LLMs under a matched protocol with a
theory-free measurement instrument.

\textbf{Validity of LLM social simulation.} LLMs as synthetic respondents trace back to
silicon sampling \citep{argyle2023out,horton2023large}; their use is under validation debate
\citep{ziems2024can,bail2024can,anthis2025llm}. Intrinsic dispositions include political stances
\citep{santurkar2023whose}, sycophancy \citep{sharma2023understanding}, simulation bias
\citep{qu2024performance}, homogenization \citep{doshi2024generative}, homophily-driven
amplification \citep{mehdizadeh2025homophily}, and skewed judgment \citep{loru2025simulation}.
We \emph{measure} such dispositions by fitting classical opinion dynamics models to theory-free measurements.

\subsection{Contributions}
\textbf{Theory-free measurement.} Opinion measurements use a psychometric rule independent of
any dynamics model, and exposure never references opinion values; fits are falsifiable
(Section~\ref{sec:prelim}).

\textbf{Cross-LLM model selection.} Classical models fitted to six LLMs under matched
conditions show that \emph{alignment drift} from a persona-assigned stance under
exposure is LLM-specific (Section~\ref{sec:results}).

\textbf{Robustness.} A no-discussion control separates social from intrinsic change; qualitative profiles
persist across population, horizon, exposure, identity, and two issues
(Section~\ref{sec:robust}, Appendix~\ref{app:controls}).

\section{Preliminaries: framework, measurement, and hypotheses}
\label{sec:prelim}

\paragraph{Discussion protocol.}
A run instantiates $n$ agents ($n{=}10$), each with a fixed one-sentence \emph{persona} and
initial \emph{stance} $x_i(0)\in[-1,1]$ on a contested policy issue
(government subsidies for electric vehicles; $x{=}-1$ ``subsidies are wasteful and EVs should
stand on their own,'' $x{=}+1$ ``aggressive public subsidy is justified and urgent''). Each of
$T$ rounds (typically $T{=}20$) has a synchronous \emph{compose} phase (every agent writes a
statement conditioned on the previous round) and a \emph{survey} phase (every agent re-answers
the instrument). Under \emph{plenary} exposure every agent hears every statement. Two optional
conditions modify only exposure or influence: \emph{selective attention} (each agent names,
from prose alone, the peers it takes seriously and hears only that subset) and a
\emph{group-identity} condition \citep{widler2025investigation} (labels tied to initial stance,
verbalized group positions, in-group weighting, label migration). The agent's own recent
opinions are resubmitted as context under a \emph{framing condition} (anchored-strict,
anchored-soft, persuadable); no opinion anchor is built into the agents.

\paragraph{Theory-free opinion measurement.}
The opinion coordinate is elicited with a fixed six-item Likert instrument (three
reverse-coded; Appendix~\ref{app:instrument}) and computed by the summated-ratings composite
\begin{equation}
x \;=\; \mathrm{clip}_{[-1,1]}\!\left(
\frac{\sum_k w_k\, s_k\,\bigl(q_k-\tfrac{L+1}{2}\bigr)}{\tfrac{L-1}{2}\,\sum_k w_k}\right),
\label{eq:composite}
\end{equation}
with grades $q_k\in\{1,\dots,7\}$, polarities $s_k\in\{-1,+1\}$ and fixed weights $w_k$;
reverse-coded items force internal consistency, so uniform agreement scores $x{=}0$.
Measurement is kept separate from theory: no opinion-dynamics model enters the measurement
path, exposure is never gated by opinion distance, and the storage layer refuses non-LLM data
to the fitting code. Only after measurement we fit a battery of classical models by one-step
least squares with BIC selection, following \citep{Helfmann2023}: persistence, DeGroot
\citep{degroot1974reaching}, Friedkin--Johnsen \citep{friedkin1990social}, bounded confidence
(BC with radius $r$) \citep{hegselmann2002opinion,deffuant2000mixing}, and a common-drift
attractor. Outcomes are summarized by the population dispersion $s(t)$ (cross-sectional s.d.\
of $\{x_i(t)\}$) and, under selective attention, by a \emph{homophily index}: the difference
between the mean opinion distance to dropped and retained peers, from the recorded
who-heard-whom data (positive $=$ stance-homophilous attention).

\paragraph{Alignment and alignment drift.}
An agent is \emph{aligned} with its persona to the extent its expressed stance tracks the
assigned stance; \emph{alignment drift} is the loss of that stance under discussion. We
measure it by the mean absolute displacement $D=\frac1n\sum_i|x_i(T)-x_i(0)|$ and the sign
retention $S=\frac1n\sum_i\mathbf{1}[\mathrm{sign}\,x_i(T)=\mathrm{sign}\,x_i(0)]$, and we
run a \emph{no-discussion} control (same personas, instrument, framing and horizon, but no
peer statements and no compose phase, own-history memory retained) to measure movement
without social input. The social component of drift is the difference between the discussion
and no-discussion values of $D$ and of the population dispersion $s(T)$. In
anchored-averaging terms, alignment strength is the weight an agent retains for its own
position. We hypothesize: \textbf{H1}~LLM discussion trajectories are described
by members of the classical model family, with model-specific parameters; \textbf{H2}~the
qualitative outcome (consensus vs.\ sustained dispersion vs.\ polarization) depends on the
underlying LLM, not only on the protocol.

\section{Experiments}
\label{sec:results}

\begin{figure}[t]
\centering
\includegraphics[width=0.90\textwidth]{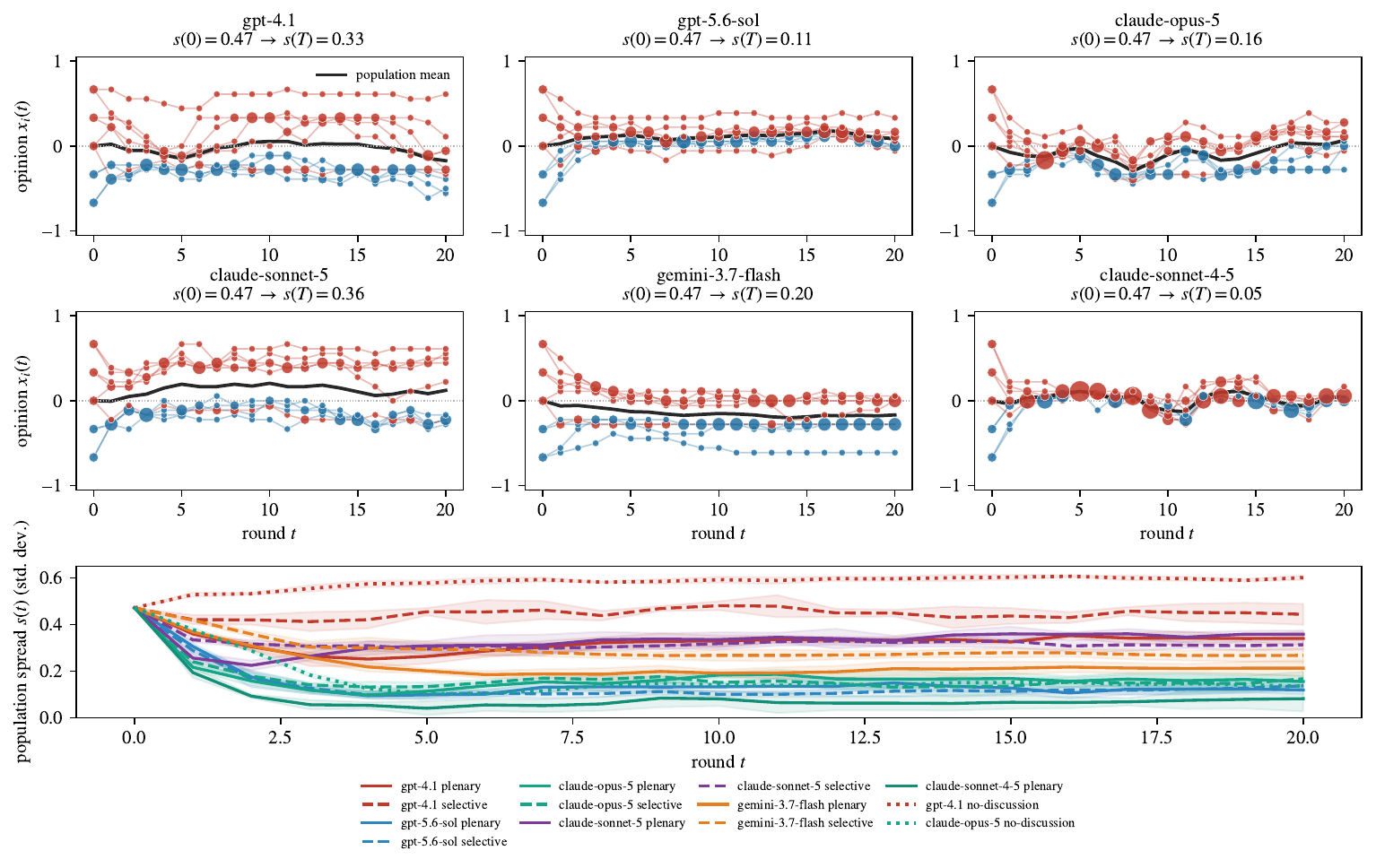}
\caption{\small Top: one representative plenary replicate per LLM (individual opinions
$x_i(t)$; color $=$ initial side; marker area $=$ number of agents sharing a value). Bottom:
ensemble dispersion $s(t)$ under plenary (solid) and selective attention (dashed); dotted
lines are the no-discussion control for \gptA{} and \opus{}. Without peers \gptA{} diverges to
$0.60$ while \opus{} collapses to $0.14$; selective attention keeps the spread high for
\gptA{} ($0.44$) and \sonnet{} ($0.31$) but not for the consensus forming LLMs.}
\label{fig:main}
\end{figure}

All runs share the measurement layer of Section~\ref{sec:prelim}; grade parsing succeeded on
$100\%$ of surveys across all reported runs. Figure~\ref{fig:main} and
Table~\ref{tab:profile} summarize the central results. 

\paragraph{Unanimous exposure: anchored averaging with LLM-specific alignment.}
Under plenary exposure from a mixed panel (initial dispersion $0.47$), every LLM population
reduces disagreement, but the degree of convergence differs strongly across LLMs
(Fig.~\ref{fig:main}, top): final dispersion after $T{=}20$ rounds ($R{=}5$ replicates) ranges
from $0.08$ (\sonnetold) and $0.12$ (\gptB) to $0.34$ (\gptA) and $0.36$ (\sonnet). Model
fitting selects a battery member for every LLM, with LLM-specific parameters
(Table~\ref{tab:fits}): Friedkin--Johnsen for \gptA{}, \gptB{} and \opus{} ($5/5$ each), the
common-drift attractor for \gem{} ($4/5$), and the pole-directed \emph{polarize} model for
\sonnet{} ($5/5$); for \sonnetold{} the leading models are indistinguishable
($\Delta\mathrm{BIC}{<}1$). Table~\ref{tab:fits} gives the fit details: evidence is strong for the
consensus-formers and moderate for \gptA{}. \textbf{H1} holds
in the weak sense that every trajectory is best described by a battery member with
LLM-specific parameters; the class itself varies by LLM.

\paragraph{Social versus intrinsic stance change.}
The plenary comparison cannot separate peer influence from repeated surveying. A
no-discussion control (same protocol, no peer statements) shows the firm population moves on its own: without peers \gptA{} \emph{diverges} ($0.47\to0.60$), whereas
\opus{} collapses ($0.47\to0.14$). Discussion therefore does not simply pull agents together: relative to this baseline it reduces dispersion for \gptA{} ($-0.26$) and \gptB{}
($-0.16$), but increases it for \sonnet{} ($+0.09$) and leaves \opus{} unchanged ($+0.02$;
Appendix~\ref{app:controls}). The committed-persona regime isolates the social component more
cleanly: without peer input all five LLMs keep every agent on its assigned side ($S{=}1.00$;
Table~\ref{tab:drift}).

\paragraph{Assigned stances: committed personas and group identity.}
With ten \emph{committed} bipolar personas, the LLMs split into drift and resistance: \gptA{} and \gem{} 
keep two separated opinion clusters ($0.75\pm0.01$, $0.71$; \gptA{}'s pro
stance group radicalizes outward, $+0.67\to+0.91$), \sonnet{} holds partially ($0.50$), while \gptB{}
and \opus{} abandon their stances under peer pressure ($0.39$, $0.18$). The no-discussion
control shows this collapse is interaction-induced: without peers all five LLMs keep every agent
on its assigned side ($S{=}1.00$), whereas discussion reduces the dispersion of \gptB{} and
\opus{} from $0.58$ and $0.41$ to $0.39$ and $0.18$, raises mean displacement from
$0.16/0.26$ to $0.33/0.51$, and moves \opus{} agents across sides ($S{:}~1.00\to0.60$); \gptA{} and \gem{} hold with or without peers (Table~\ref{tab:drift}).
An imposed \emph{group-identity} condition is the strongest lever: \gptA{} polarizes outright
(dispersion $0.47\to0.58\pm0.04$; Table~\ref{tab:profile}), whereas all other LLMs contract or
hold. This is an \emph{induced} regime: from a neutral start no population polarizes
spontaneously, and amplification maintains but never creates disagreement
(Appendix~\ref{app:controls}).

\paragraph{Selective attention: emergent homophily.}
Under selective attention, agents hear only the peers they name. Only \gptA{} (and
more weakly \sonnet{}) exhibits stance-homophilous attention (homophily index
$+0.485\pm0.032$ and $+0.327\pm0.024$), which holds two stance-separated clusters in place
(dispersion $0.44$ vs.\ $0.34$ plenary); the consensus-forming LLMs 
converge under both conditions (homophily $\approx0$). \gptB{} is the sharp opposite of \gptA{}
(homophily $-0.039\pm0.008$): the effect is a property of the agent/LLM, not of the protocol.

\paragraph{Robustness.}
\label{sec:robust}
A second issue (four-day week; six LLMs, $R{=}3$--$5$) preserves the cross-LLM ordering (Spearman $\rho{=}0.83$, Pearson $r{=}0.95$) and the
Friedkin--Johnsen class for the averaging models, while polarize/attractor labels can change
(\sonnet: polarize$\to$FJ; \gem: attractor$\to$polarize; Appendix~\ref{app:topics}). At
$n{=}20$, $T{=}40$ the qualitative profiles persist for all five re-run LLMs (\gptA{}'s
selective radius $+0.34$ and identity split gap $\approx1.0$; the other four remain
consensus-formers), but the exact BIC class shifts in some cells (\gptA{} attractor 3/FJ 2;
\gptB{} polarize 3, sub-threshold). We therefore claim robustness of the profile and ordering,
but not of the exact model at every scale. \textbf{H2 is supported}.

\begin{table}[t]
\centering\footnotesize
\setlength{\tabcolsep}{3pt}
\renewcommand{\arraystretch}{0.90}
\begin{tabular}{@{}lccccc@{}}
\toprule
LLM & Plenary $s(T)$ & Selective $s(T)$ & Homophily & Identity $s(T)$ & best model (reps) \\
\midrule
\gptA          & $0.34$ & $0.44$ & $\mathbf{+0.49}$ & $\mathbf{0.58\pm0.04}$ & FJ ($5/5$) \\
\sonnet        & $0.36$ & $0.31$ & $+0.33$ & $0.47\pm0.02$ & polarize ($5/5$) \\
\gem           & $0.21$ & $0.27$ & $+0.17$ & $0.23\pm0.05$ & attractor ($4/5$) \\
\opus          & $0.16$ & $0.17$ & $+0.00$ & $0.26\pm0.05$ & FJ ($5/5$) \\
\gptB          & $0.12$ & $0.14$ & $-0.04$ & $0.19\pm0.03$ & FJ ($5/5$) \\
\sonnetold     & $0.08$ & $0.12$ & $-0.01$ & $0.14\pm0.02$ & unresolved \\
\bottomrule
\end{tabular}
\caption{\small Behavioural profiles on the electric-vehicle-subsidy issue ($n{=}10$, $T{=}20$, $R{=}5$;
initial dispersion $0.47$). $s(T)$: final population dispersion; homophily: stance-correlation
of selective attention; identity: dispersion under the group-identity condition; best
BIC-selected opinion model (selection frequency). Full fits are in Table~\ref{tab:fits}.}
\label{tab:profile}
\end{table}

\section{Discussion and Limitations}
Our results quantify how LLM agents bring their own dispositions into social simulations. The same protocol produces consensus in some LLMs and persistent or growing division in
others; fits convert these differences into interpretable quantities: anchoring strength, homophily of selective attention, and
pole-directed pull. The observed dynamics combine intrinsic movement with \emph{alignment drift},
whose magnitude and mechanism differ by LLM. For LLM-based simulation, findings
from a single LLM may conflate its alignment profile with properties of the simulated process;
the choice of LLM must be reported and ideally varied, as one would report sample
demographics. Transfer across issues indicates stable LLM properties: the LLM becomes a
variable to ablate and control.

\paragraph{Limitations.}
Our measurement compresses each stance to one axis on two policy issues; populations are
homogeneous (one LLM per society); experiments use English prompts, proprietary models, and
$3$--$5$ replicates; we pin model IDs and dates and release code/data; snapshots can change. Polarization required committed personas or identity, so our claims concern maintaining and
amplifying divisions, not creating them. Both issues are economic-policy debates;
non-economic domains remain untested. The
no-discussion control removes both peer input and statement composition, so it measures change in the absence of discussion but does not isolate peer exposure. The single axis omits
multi-dimensional value trade-offs and emotional dynamics, although the six items are
internally consistent (Cronbach $\alpha{=}0.88$--$0.97$; pro-vs-reverse $r{=}-0.83$ to
$-0.95$; straight-lining $1$--$1.5\%$). Jump filtering, prompts, and interaction geometry may shift profiles.

\bibliographystyle{plainnat}
\bibliography{references}

\clearpage
\appendix

\section{Survey instrument}
\label{app:instrument}
Each round, agents rate six declarative statements on an integer scale $1$ (strongly
disagree) to $7$ (strongly agree); three are worded toward each pole (reverse-coded items
detect acquiescence). EV-subsidy issue (polarity $s_k$ in parentheses):
(Q1, $+$) government subsidies are the fastest realistic way to cut transport carbon
emissions; (Q2, $-$) EV subsidies mainly benefit wealthier buyers who would have bought an
EV anyway; (Q3, $+$) adequate charging infrastructure will not be built fast enough without
government funding; (Q4, $-$) the market, not taxpayer money, should decide how quickly EVs
replace petrol cars; (Q5, $+$) the health and climate benefits of EV subsidies outweigh
their cost to taxpayers; (Q6, $-$) public money spent on EV subsidies would do more good if
spent on other priorities. The four-day-week instrument follows the same construction
(three pro / three con items on trials and wellbeing vs.\ competitiveness, coverage, and
affordability), and the composite Eq.~\eqref{eq:composite} is identical. Across all plenary
surveys the six items are internally consistent (Cronbach $\alpha{=}0.88$--$0.97$; pro- vs.\
reverse-coded subscale $r{=}-0.83$ to $-0.95$; within-agent item s.d.\ $0.9$--$1.1$;
straight-lining $1$--$1.5\%$), so the scalar is a reliable composite rather than a single
Likert answer.

\section{Alternative topics}
\label{app:topics}
 Additional, unrelated issues were also tested, e.g., whether governments and employers should move to a standard four-day (32-hour) work week at full pay; each uses the same survey form with adapted questions and an opinion
 axis running from $-1$ (``the five-day week should stay; a shorter week is impractical and costly'') to $+1$ (``a four-day week should become the norm, improving well-being and
 productivity''), so a backbone's dynamics can be compared across topics under the same measurement procedure. The plenary ordering transfers (Spearman $\rho{=}0.83$, Pearson $r{=}0.95$ across the six LLMs); the averaging models stay Friedkin--Johnsen, while \sonnet{} switches from polarize to FJ and \gem{} from attractor to polarize (Table~\ref{tab:topics}).

\begin{table}[h]\centering\footnotesize
\setlength{\tabcolsep}{4pt}
\begin{tabular}{@{}lcccc@{}}
\toprule
LLM & EV $s(T)$ & 4-day $s(T)$ & $\Delta$ & 4-day model \\
\midrule
\gptA      & $0.34$ & $0.21$ & $-0.13$ & FJ ($5/5$) \\
\sonnet    & $0.36$ & $0.25$ & $-0.11$ & FJ ($3/3$) \\
\gem       & $0.21$ & $0.16$ & $-0.05$ & polarize ($3/3$) \\
\opus      & $0.16$ & $0.11$ & $-0.05$ & FJ ($2/3$) \\
\gptB      & $0.12$ & $0.13$ & $+0.01$ & FJ ($3/3$) \\
\sonnetold & $0.08$ & $0.12$ & $+0.04$ & FJ ($5/5$) \\
\bottomrule
\end{tabular}
\caption{\small Second-topic replication (four-day work week, plenary, $n{=}10$, $T{=}20$;
$R{=}5$ for \gptA{}, \sonnetold{} and $R{=}3$ otherwise). The cross-LLM ordering is
preserved; FJ is stable for the averaging models, while \sonnet{} switches from polarize to FJ
and \gem{} from attractor to polarize.}
\label{tab:topics}
\end{table}

\section{Personas}
\label{app:personas}
The firm-but-uncommitted roster used for the cross-LLM grid consists of ten one-sentence
personas with initial opinions evenly spaced on $[-0.8,+0.8]$ (e.g.\ ``a rural mechanic
skeptical of EV practicality and cost'', \ldots, ``an EV early-adopter and tech optimist''),
so that every population starts with an identical stance distribution. The committed roster
comprises five strongly pro and five strongly con personas seeded at interior stances
$|x_i(0)|\le0.7$ (e.g.\ ``a climate scientist convinced rapid EV subsidies are a moral
necessity'' vs.\ ``a taxpayer-union head opposed to large public spending on principle'').
The neutral roster seeds all agents at $x_i(0)=0$ with side-agnostic personas for the
spontaneous-polarization control. Full verbatim lists are available from the authors.

\section{Model fit quality}
\label{app:fits}
Table~\ref{tab:fits} reports the information-criterion margins, fitted parameters and
residual errors behind the model selections. $\Delta$BIC is given relative to the
persistence (no-change) null and to the runner-up; one-step RMSE is the error in predicting
$x(t{+}1)$ from the true $x(t)$, and free-run RMSE is the error of a simulation launched from
$x(0)$ alone, with the persistence free-run error in parentheses. The battery recovers known generating models and parameters on synthetic data, so these
margins reflect the data rather than an estimator artefact.

\begin{table}[h]\centering\scriptsize
\setlength{\tabcolsep}{1pt}
\begin{tabular}{@{}lcccccc@{}}
\toprule
LLM & model (reps) & $\Delta$BIC$_{\mathrm{null}}$ & $\Delta$BIC$_{\mathrm{next}}$ & parameters & RMSE$_1$ & free (pers.) \\
\midrule
\gptA      & FJ ($5/5$)       & $7.1$   & $3.0$  & $\beta{=}0.92{\pm}0.01,\lambda{=}0.92{\pm}0.02$ & $0.093$ & $0.195$ ($0.260$) \\
\sonnet    & polarize ($5/5$) & $14.5$  & $3.7$  & $\beta{=}0.77{\pm}0.08,\ \lambda{=}0.87{\pm}0.03,a{=}0.83$ & $0.087$ & $0.235$ ($0.264$) \\
\gem       & attractor ($4/5$) & $38.8$ & $5.9$  & $\beta{=}1.00{\pm}0.00,\ \lambda{=}0.90{\pm}0.00,c{=}-0.29$ & $0.054$ & $0.140$ ($0.342$) \\
\opus      & FJ ($5/5$)       & $73.7$  & $18.5$ & $\beta{=}0.60{\pm}0.03,\lambda{=}0.87{\pm}0.03$ & $0.085$ & $0.127$ ($0.353$) \\
\gptB      & FJ ($5/5$)       & $88.7$  & $13.6$ & $\beta{=}0.70{\pm}0.01,\lambda{=}0.94{\pm}0.01$ & $0.059$ & $0.121$ ($0.375$) \\
\sonnetold & unresolved       & $115.4$ & $0.7$  & --- & $0.088$ & $0.144$ ($0.446$) \\
\bottomrule
\end{tabular}
\caption{\small Model-fit quality on the plenary EV trajectories (medians over replicates).
$\Delta$BIC$_{\mathrm{null}}$: improvement over the persistence null;
$\Delta$BIC$_{\mathrm{next}}$: margin over the runner-up. Parameters are the modal-winner
estimates ($\beta,\lambda$: mean$\pm$s.d.\ over its replicates). RMSE$_1$: one-step; free: free-running from $x(0)$;
persistence free-run in parentheses. For \sonnetold{} the leading models are statistically
indistinguishable ($\Delta\mathrm{BIC}_{\mathrm{next}}{<}1$).}
\label{tab:fits}
\end{table}

\section{Additional controls}
\label{app:controls}
\textbf{Spontaneous polarization.} From a neutral start ($x_i(0)=0$, side-agnostic personas,
$T{=}30$) all populations converge under both conditions (final dispersion $0.031$ plenary, $0.027$
selective; polarized each), and the homophily signal fails to emerge ($+0.002$ vs.\
$+0.485$ when an initial spread is present): selective attention maintains and structures
pre-existing disagreement but does not create it.
\textbf{Amplified conditions.} A very-strict attention variant (keep only the single most
compelling peer) and a strong biased-assimilation influence condition polarize none of the tested LLM populations
($0/5$ in all cells); \gptA{} retains a wide dispersion and a positive homophily signal under
both, i.e.\ amplification maintains but does not grow disagreement.
\textbf{No-discussion baseline (firm roster).} Without peer input the firm-but-uncommitted
population is not stable either: \gptA{} diverges ($0.47\to0.60$), whereas \opus{} collapses
($0.47\to0.14$); discussion reduces dispersion relative to this baseline for \gptA{}
($-0.26$), \gptB{} ($-0.16$), \gem{} ($-0.14$) and \sonnetold{} ($-0.08$), but increases it
for \sonnet{} ($+0.09$) and leaves \opus{} unchanged ($+0.02$).
\textbf{Scale.} At $n{=}20$, $T{=}40$ all five re-run LLMs preserve the qualitative
profiles and mechanism: \gptA{} keeps the selective-attention radius ($+0.34$) and the identity split
($\approx1.0$), while \gptB{}, \opus{}, \gem{} and \sonnet{} remain consensus-formers; the exact BIC class shifts in some cells (\gptA{} attractor 3/FJ 2;
\gptB{} polarize 3, sub-threshold).

\begin{table}[h]
\centering\footnotesize
\setlength{\tabcolsep}{4pt}
\begin{tabular}{@{}lcccccc@{}}
\toprule
LLM & $s(T)$ disc. & $s(T)$ no-disc. & $\Delta s(T)$ & $D$ disc. & $D$ no-disc. & $S$ disc. \\
\midrule
\gptA      & $0.75$ & $0.85$ & $-0.10$ & $0.16$ & $0.19$ & $1.00$ \\
\gem       & $0.71$ & $0.67$ & $+0.05$ & $0.17$ & $0.11$ & $1.00$ \\
\sonnet    & $0.50$ & $0.59$ & $-0.09$ & $0.24$ & $0.08$ & $0.97$ \\
\gptB      & $0.39$ & $0.58$ & $-0.19$ & $0.33$ & $0.16$ & $0.90$ \\
\opus      & $0.18$ & $0.41$ & $-0.23$ & $0.51$ & $0.26$ & $0.60$ \\
\bottomrule
\end{tabular}
\caption{\small Committed-persona regime ($n{=}10$, $T{=}20$; discussion $R{=}5$ for \gptA{},
$R{=}3$ otherwise; no-discussion $R{=}3$): final dispersion $s(T)$, mean per-agent displacement
$D{=}\frac1n\sum_i|x_i(T)-x_i(0)|$, and sign retention $S$, with peer discussion versus the
no-discussion control (no peer statements, no compose phase, own-history memory retained).
Without peer input all five LLMs keep every agent on its assigned side ($S{=}1.00$);
discussion collapses \gptB{} and \opus{}, with side crossings in both populations,
more frequently for \opus{}, while \gptA{} and \gem{} hold.}
\label{tab:drift}
\end{table}

\paragraph{Per-agent displacement.}
The committed-persona collapse is broad-based rather than driven by a few agents. Under
discussion, \opus{}'s median per-agent displacement is $0.53$ (IQR $0.33$--$0.67$), half of
its agents move more than $0.5$ units and $20\%$ cross the midpoint; without peers the median
is $0.22$ (IQR $0.17$--$0.33$) and no agent moves more than $0.5$ or crosses. For \gptB{} the
values are $0.36$ (IQR $0.06$--$0.51$; $23\%$ above $0.5$, $10\%$ crossing) versus $0.11$
($0.06$--$0.28$; $0\%$). \gptA{} stays small under both conditions (median $0.19$ discussion,
$0.28$ no-discussion; no agent above $0.5$).

\end{document}